\documentclass[conference]{IEEEtran}

\usepackage{amsmath}
\usepackage{amssymb}
\usepackage{graphicx}
\usepackage{cite}
\usepackage{booktabs}
\usepackage{url}

\begin{document}

\title{Quantum Generative Models for Computational Fluid Dynamics:
A First Exploration of Latent Space Learning in Lattice Boltzmann Simulations}

\author{
\IEEEauthorblockN{Achraf Hsain}
\IEEEauthorblockA{Al Akhawayn University\\
a.hsain@aui.ma}
\and
\IEEEauthorblockN{Fouad Mohammed Abbou}
\IEEEauthorblockA{Al Akhawayn University\\
f.abbou@aui.ma}
}

\maketitle

\begin{abstract}
This paper presents the first application of quantum generative models to learned latent space representations of computational fluid dynamics (CFD) data. While recent work has explored quantum models for learning statistical properties of fluid systems, the combination of discrete latent space compression with quantum generative sampling for CFD remains unexplored. We develop a GPU-accelerated Lattice Boltzmann Method (LBM) simulator to generate fluid vorticity fields, which are compressed into a discrete 7-dimensional latent space using a Vector Quantized Variational Autoencoder (VQ-VAE). The central contribution is a comparative analysis of quantum and classical generative approaches for modeling this physics-derived latent distribution: we evaluate a Quantum Circuit Born Machine (QCBM) and Quantum Generative Adversarial Network (QGAN) against a classical Long Short-Term Memory (LSTM) baseline. Under our experimental conditions, both quantum models produced samples with lower average minimum distances to the true distribution compared to the LSTM, with the QCBM achieving the most favorable metrics. This work provides: (1)~a complete open-source pipeline bridging CFD simulation and quantum machine learning, (2)~the first empirical study of quantum generative modeling on compressed latent representations of physics simulations, and (3)~a foundation for future rigorous investigation at this intersection.
\end{abstract}

\begin{IEEEkeywords}
Quantum Machine Learning, Computational Fluid Dynamics, Generative Models, Lattice Boltzmann Method, Vector Quantized Variational Autoencoder, Quantum Circuit Born Machine, Quantum Generative Adversarial Network, Latent Space Modeling, Physics-Informed Machine Learning
\end{IEEEkeywords}

\section{Introduction}
Computational Fluid Dynamics (CFD) plays a crucial role in understanding and predicting the behavior of fluid systems across a wide range of scientific and engineering disciplines, including aerospace, automotive, environmental science, and energy \cite{1, 2}. Traditional methods for solving fluid dynamics problems, such as those based on the Navier-Stokes equations \cite{22, 23, 24}, often involve significant computational resources, particularly for complex or turbulent flows \cite{3}. The increasing complexity of simulations and the resulting high-dimensional data necessitate the development of more efficient methods for data representation, analysis, and generation.

Generative models have emerged as powerful tools for learning the underlying probability distributions of complex datasets and generating novel, synthetic data samples that resemble the original data \cite{4, 5, 6}. These models can capture intricate patterns and structures within data, offering potential benefits for tasks such as data augmentation, anomaly detection, and understanding complex system dynamics.

Dimensionality reduction techniques are often employed to handle the high-dimensional output of simulations. Autoencoders and Variational Autoencoders (VAEs) are neural network architectures capable of learning compressed, lower-dimensional latent space representations of data \cite{6, 7, 8}. VAEs provide a probabilistic framework for encoding data into a continuous latent space, enabling the generation of new data by sampling from this learned distribution \cite{6, 11}. Vector Quantized Variational Autoencoders (VQ-VAEs) extend this by learning a discrete latent space represented by a codebook of embeddings, which can be particularly advantageous for tasks requiring discrete inputs or for integration with certain computational paradigms \cite{13}.

Concurrently, the field of quantum computing is rapidly advancing, demonstrating potential advantages for specific computational problems, including certain machine learning tasks \cite{14}. Quantum machine learning algorithms, such as Quantum Circuit Born Machines (QCBMs) and Quantum Generative Adversarial Networks (QGANs), leverage the principles of quantum mechanics to potentially model complex probability distributions more effectively than classical methods \cite{14, 15, 16, 17}. 

Despite growing interest in both quantum machine learning and data-driven approaches to CFD, the intersection of these fields---specifically, applying quantum generative models to learn distributions over compressed representations of physics simulation data---remains largely unexplored. Recent work has begun investigating quantum models for learning statistical properties of chaotic fluid systems \cite{31}, demonstrating that QCBMs can capture invariant measures of turbulent flows. However, the application of quantum generative models to explicitly learned latent space representations---particularly discrete codebook representations from architectures like VQ-VAE---has not been explored for CFD data. This paper addresses this gap by presenting what is, to our knowledge, the first investigation of quantum generative models applied to compressed latent representations of computational fluid dynamics simulations.

We propose an end-to-end approach that combines: (1)~a GPU-accelerated Lattice Boltzmann Method simulator for fluid dynamics data generation, (2)~a VQ-VAE for learning a discrete latent representation of fluid vorticity, and (3)~a comparative study of classical LSTM networks, QCBMs, and QGANs for modeling and sampling from this latent space prior distribution.

\subsection{Scope and Contributions}
This work is an exploratory study rather than a definitive benchmark. We do not claim quantum advantage---our quantum models run on classical simulators, and the comparison involves a single classical baseline under specific experimental conditions. The primary contributions are:

\begin{itemize}
    \item \textbf{Novel application domain}: The first application of quantum generative models (QCBM and QGAN) to learned latent space representations of CFD data. While prior work has explored quantum models for fluid statistics \cite{31}, the combination of discrete latent compression (VQ-VAE) with quantum generative sampling for physics simulations is new.
    
    \item \textbf{Open-source infrastructure}: A complete, reproducible pipeline connecting LBM fluid simulation, VQ-VAE latent space compression, and quantum/classical generative modeling, providing a foundation for future research in this domain.
    
    \item \textbf{Empirical observations}: Comparative results between quantum and classical generative approaches under controlled conditions, with detailed documentation of architectures, hyperparameters, and training procedures to enable reproducibility and extension.
    
    \item \textbf{Research directions}: Discussion of observed performance differences, potential explanations, and specific hypotheses for more rigorous future investigation.
\end{itemize}

The remainder of this paper is organized as follows. Section~II provides background on the theoretical concepts underpinning this research. Section~III states the problem and objectives. Section~IV describes the methodology employed. Section~V presents the experimental results and analysis. Section~VI discusses the implications and limitations of our findings. Section~VII outlines future work, and Section~VIII concludes the paper.

\section{Literature Review}
This section provides an overview of the foundational concepts and prior research relevant to this work, spanning computational fluid dynamics, generative modeling, and quantum machine learning.

\subsection{Fluid Dynamics and Computational Fluid Dynamics}
Fluid mechanics is a fundamental branch of physics concerned with the motion of fluids (liquids, gases, and plasmas) and the forces acting upon them. The behavior of Newtonian fluids is primarily governed by the Navier-Stokes equations, a set of partial differential equations describing the conservation of momentum and mass \cite{22, 23, 24}. These equations are notoriously difficult to solve analytically, especially for complex geometries or turbulent flow regimes. Turbulence, characterized by chaotic, unpredictable fluid motion, remains one of the most challenging problems in classical physics and engineering \cite{3}.

Computational Fluid Dynamics (CFD) emerged as a discipline to tackle these challenges by employing numerical methods and algorithms to simulate fluid flows \cite{1, 2}. CFD has become an indispensable tool in diverse fields, enabling the design and analysis of aircraft wings, optimization of automotive aerodynamics, prediction of weather patterns, and simulation of blood flow in biological systems. Common numerical approaches include Finite Difference Methods, Finite Volume Methods, and Finite Element Methods.

A key dimensionless parameter in fluid dynamics is the Reynolds number (Re), which represents the ratio of inertial forces to viscous forces within a fluid \cite{24}. It is a critical indicator of flow regime; low Reynolds numbers typically correspond to laminar flow, while high Reynolds numbers are associated with turbulent flow. Accurate simulation of high Reynolds number flows remains computationally intensive, driving the need for more efficient simulation techniques and data analysis methods.

\subsection{Lattice Boltzmann Method}
The Lattice Boltzmann Method (LBM) is an alternative CFD approach that has gained significant popularity due to its conceptual simplicity, computational efficiency, and suitability for parallel implementation \cite{18}. Unlike traditional CFD methods that directly discretize macroscopic conservation equations, LBM is based on a simplified kinetic model. It simulates the statistical behavior of fluid particles on a discrete lattice grid, tracking the evolution of particle distribution functions. Macroscopic properties such as density, velocity, and pressure are then derived from moments of these distribution functions.

The LBM typically involves two main steps: collision and streaming. In the collision step, particle distribution functions at each lattice node relax towards a local equilibrium distribution, often modeled by the Bhatnagar-Gross-Krook (BGK) approximation. The streaming step involves the movement of particles to neighboring lattice nodes according to their discrete velocities.

The lattice structure is defined by the DnQm notation, where `n' is the number of spatial dimensions and `m' is the number of discrete velocity vectors. The D2Q9 model, used in this work, is well-suited for simulating incompressible or nearly incompressible flows \cite{18}. LBM's inherent locality and explicit time-stepping make it highly amenable to parallel processing, particularly on GPUs, enabling significant acceleration for large-scale simulations \cite{19}. Boundary conditions, such as no-slip boundaries and inflow/outflow conditions, are implemented to define the interaction of the fluid with solid objects and domain boundaries \cite{20, 21}.

\subsection{Autoencoders and Variational Autoencoders}
Dimensionality reduction is a critical step in analyzing and processing high-dimensional data, including CFD simulation outputs. Autoencoders are neural networks designed to learn efficient data encodings \cite{7, 8}. An autoencoder consists of an encoder network that maps input data to a lower-dimensional latent space representation and a decoder network that reconstructs the original data from this latent representation.

Variational Autoencoders (VAEs) extend the autoencoder framework by introducing a probabilistic perspective \cite{6}. Instead of mapping an input to a fixed point in the latent space, the encoder maps it to parameters of a probability distribution in the latent space. This probabilistic encoding allows VAEs to generate new data samples by sampling from the learned latent distributions. VAEs are trained by optimizing the Evidence Lower Bound (ELBO), which incorporates both reconstruction loss and a regularization term encouraging the latent distributions to be close to a prior distribution \cite{6, 11, 26}.

\subsection{Vector Quantized Variational Autoencoders}
While VAEs learn a continuous latent space, some applications benefit from discrete latent representations. The Vector Quantized Variational Autoencoder (VQ-VAE) modifies the standard VAE by replacing the continuous latent space with a discrete codebook of learned embedding vectors \cite{13}. The encoder outputs a representation that is ``quantized'' by finding the closest embedding vector in the codebook. This discrete codebook vector is then passed to the decoder.

Training a VQ-VAE involves minimizing a loss function with three components: reconstruction loss, codebook loss (updating embeddings based on encoder outputs), and commitment loss (encouraging encoder output to stay close to the chosen codebook vector). VQ-VAEs have been successfully applied in image generation, video prediction, and speech synthesis. Their capacity to produce discrete latent codes makes them particularly relevant for integration with models that require discrete inputs, such as certain quantum algorithms \cite{13}.

\subsection{Quantum Computing Fundamentals}
Quantum computing leverages the principles of quantum mechanics to perform computations. Unlike classical bits that represent either 0 or 1, quantum bits (qubits) can exist in a superposition of both states simultaneously. The state of a system of $n$ qubits exists in a $2^n$-dimensional complex vector space, allowing for the representation of exponentially large state spaces.

Quantum gates are unitary operations that manipulate qubit states. Sequences of quantum gates form quantum circuits, which perform the desired computation. Entanglement, where the states of two or more qubits become correlated in a way that cannot be described classically, is a crucial resource for many quantum algorithms \cite{14}. While large-scale fault-tolerant quantum computers are still under development, noisy intermediate-scale quantum (NISQ) devices are currently available, and classical simulators can emulate quantum circuits for a limited number of qubits.

\subsection{Quantum Generative Models}
Research has explored quantum generative models that leverage quantum systems to represent and manipulate high-dimensional probability distributions. These models aim to learn a target probability distribution and generate samples from it using quantum circuits.

The Quantum Circuit Born Machine (QCBM) uses a parameterized quantum circuit to define a probability distribution over computational basis states \cite{15}. The circuit parameters are trained to minimize a distance measure between the distribution sampled from the quantum circuit and the target data distribution. Training typically involves optimizing parameters using classical techniques, guided by a loss function computed from circuit measurements. The Maximum Mean Discrepancy (MMD) is a commonly used loss function, providing a kernel-based measure of the distance between two distributions \cite{27}:
\begin{equation}
\text{MMD}^2(p,q) = \mathbb{E}_{p}[k(x,x')] - 2\mathbb{E}_{p,q}[k(x,y)] + \mathbb{E}_{q}[k(y,y')]
\end{equation}
where $\mathbb{E}$ denotes expected value and the Radial Basis Function kernel is a popular choice for $k$ \cite{27}.

Quantum Generative Adversarial Networks (QGANs) adapt the classical GAN framework to the quantum domain \cite{16, 17}. A QGAN typically consists of a quantum generator that produces quantum states (or classical samples derived from them) and a discriminator that distinguishes between real and generated samples. The generator and discriminator are trained adversarially \cite{5, 16, 17}.

\subsection{Quantum Models for Physics Simulations}
The application of quantum machine learning to physics simulation data is an emerging research area. In high-energy physics, quantum GANs have been applied to particle detector simulations and calorimeter response modeling \cite{14}. More recently, quantum-informed approaches have been explored for chaotic dynamical systems, including fluid flows governed by the Navier-Stokes equations \cite{31}. This work demonstrated that QCBMs can learn statistical invariant measures of turbulent systems and provide regularizing priors for classical surrogate models. However, this approach operates on raw flow statistics rather than learned latent representations. The combination of explicit latent space compression (such as through autoencoders) with quantum generative sampling for CFD data has not been previously explored, representing the gap this work addresses.

\subsection{Relevant Classical Techniques}
Long Short-Term Memory (LSTM) networks are recurrent neural networks particularly well-suited for modeling sequential data \cite{25}. LSTMs utilize gating mechanisms to mitigate the vanishing gradient problem, allowing them to learn long-range dependencies. When training autoregressive models like LSTMs for sequence generation, teacher forcing---feeding the true previous output as input at the current time step---can stabilize and accelerate training \cite{29, 30}.

For comparative analysis, t-distributed Stochastic Neighbor Embedding (t-SNE) is a non-linear dimensionality reduction technique for visualizing high-dimensional data \cite{28}. Principal Component Analysis (PCA) provides a linear dimensionality reduction perspective \cite{8}. Metrics such as average minimum distance and nearest neighbor analysis provide quantitative measures for comparing how well generated samples match a reference distribution \cite{26, 27}.

\section{Problem Statement and Objectives}
Traditional CFD simulations, particularly for complex phenomena like turbulent flows at high Reynolds numbers, are computationally intensive and generate vast amounts of high-dimensional data \cite{3}. Analyzing, storing, and leveraging this data effectively poses significant challenges. While dimensionality reduction techniques and classical generative models offer potential solutions for data compression and synthesis, their application to the intricate patterns of fluid dynamics warrants further investigation.

Furthermore, quantum machine learning algorithms hold promise for modeling complex probability distributions that may be difficult for classical counterparts \cite{14}. However, the practical utility and performance of current quantum generative models, especially when applied to real-world scientific data like fluid dynamics, remain active areas of research. Recent work has demonstrated that quantum models can learn statistical properties of chaotic fluid systems \cite{31}, but the application of quantum generative models to explicitly learned latent space representations of physics simulation data has not been explored.

The core problem addressed in this research is the need for efficient and effective methods to model and generate representative data from the complex, high-dimensional outputs of fluid dynamics simulations. Specifically, we aim to explore how quantum generative models compare to classical methods in learning and sampling from a compressed, discrete latent space representation of fluid vorticity.

Based on this problem statement, the primary objectives of this study are:

\begin{itemize}
    \item To develop a robust computational framework capable of generating physically accurate fluid dynamics data using a GPU-accelerated LBM simulator.
    \item To apply a VQ-VAE to learn a highly compressed and discrete latent space representation of the generated fluid vorticity data.
    \item To train and evaluate classical and quantum generative models on the learned discrete latent space to model its underlying prior distribution.
    \item To conduct a comparative analysis of the trained classical and quantum generative models using both visualization techniques \cite{28} and quantitative metrics \cite{26, 27}.
    \item To document observations and generate hypotheses regarding the relative performance of these approaches, identifying directions for future rigorous investigation.
\end{itemize}

\section{Methodology}
This section details the computational framework and experimental procedures employed in this research. The methodology follows a modular approach, allowing for independent development and testing of each component. All experiments were performed using an NVIDIA RTX 4060 GPU, implemented primarily in Python using PyTorch, PennyLane, Scikit-learn, NumPy, Open3D, Matplotlib, and PyVista.

\subsection{LBM Fluid Simulator and Data Generation}
Fluid dynamics data was generated using a custom GPU-accelerated LBM simulator implemented in Python \cite{19}. The simulator was capable of both 2D and 3D simulations, although the primary dataset was generated using the 2D implementation.

The simulation framework managed execution flow and configuration of parameters such as grid dimensions, fluid properties, and object geometry. Object geometry was incorporated via a masking process. For 2D simulations, object shapes were defined from image data to create solid boundaries. Visualization capabilities were integrated to render flow fields such as velocity, vorticity, and density.

The core LBM simulation logic implemented the fundamental steps on GPU:

\begin{itemize}
\item \textbf{Inflow}: Sets velocity boundary conditions at the inlet region \cite{20}.
\item \textbf{Compute Equilibrium}: Calculates local equilibrium distribution functions based on macroscopic fluid properties, appropriate for the D2Q9 scheme \cite{18}.
\item \textbf{Collide}: Relaxes particle distribution functions towards local equilibrium, governed by the relaxation parameter ($\tau$) related to fluid viscosity \cite{18}.
\item \textbf{Bounce Back}: Implements no-slip boundary conditions at solid boundaries \cite{21}.
\item \textbf{Propagate}: Performs the streaming step, moving particle distribution functions to adjacent nodes \cite{18}.
\end{itemize}

The simulator's physical accuracy was benchmarked against established CFD results \cite{18}. The dataset used for training consisted of 2D vorticity profiles from a simulation of flow around a cylinder with radius 16 pixels, inlet velocity 0.1 Mach, and Reynolds number 500. Each snapshot represented the vorticity field as a $256 \times 64$ grid. A total of 1999 simulation snapshots were collected.

\begin{figure}[t]
  \centering
  \includegraphics[width=0.35\textwidth]{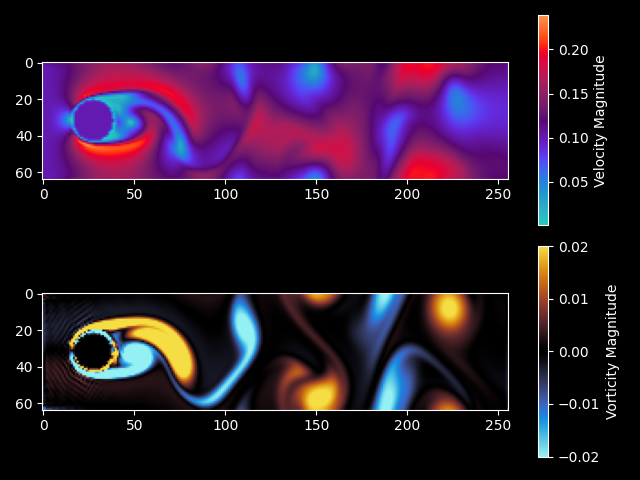}
  \caption{Snapshot from the LBM-generated dataset showing vorticity field for flow around a cylinder at Re=500.}
  \label{fig:LBM}
\end{figure}

\subsection{VQ-VAE Latent Space Learning}
To obtain a compressed, discrete latent space representation, a VQ-VAE model was employed \cite{13}. The architecture consists of three main components:

\begin{itemize}
    \item \textbf{Image Encoder}: Processes the $256 \times 64$ input vorticity grid through 2D convolutional layers with ReLU activation and batch normalization, progressively reducing spatial dimensions while increasing feature maps until obtaining a $256 \times 16 \times 4$ feature space.
    
    \item \textbf{Quantized Encoder-Decoder}: Uses MLPs to map the feature space to a 2048-dimensional vector, then to a 7-dimensional vector representing the continuous latent space output. This vector is quantized by finding the closest embedding in a discrete codebook containing 128 codewords, each with dimension 7. A separate MLP maps the selected codeword back to a 2048-dimensional vector for decoding.
    
    \item \textbf{Image Decoder}: Mirrors the encoder using transposed convolutional layers to reconstruct the original $256 \times 64$ grid.
\end{itemize}

The VQ-VAE loss function is:
\begin{equation}
\mathcal{L}_{\text{VQ-VAE}} =
\underbrace{\| x - \hat{x} \|_2^2}_{\text{Reconstruction}} +
\underbrace{\| \text{sg}[z_e(x)] - e \|_2^2}_{\text{Codebook}} +
\beta \underbrace{\| z_e(x) - \text{sg}[e] \|_2^2}_{\text{Commitment}}
\end{equation}
where $x$ is the input, $\hat{x}$ is the reconstruction, $z_e(x)$ is the encoder output, $e$ is the selected codebook embedding, $\text{sg}[\cdot]$ denotes the stop-gradient operator, and $\beta$ is the commitment loss weight.

The model was trained using Adam optimizer with learning rate 0.0005 for 50 epochs, batch size 32, and $\beta=0.2$. Straight-through estimation enabled gradient flow through the non-differentiable quantization step \cite{13}.

\begin{figure}[t]
  \centering
  \includegraphics[width=0.5\textwidth]{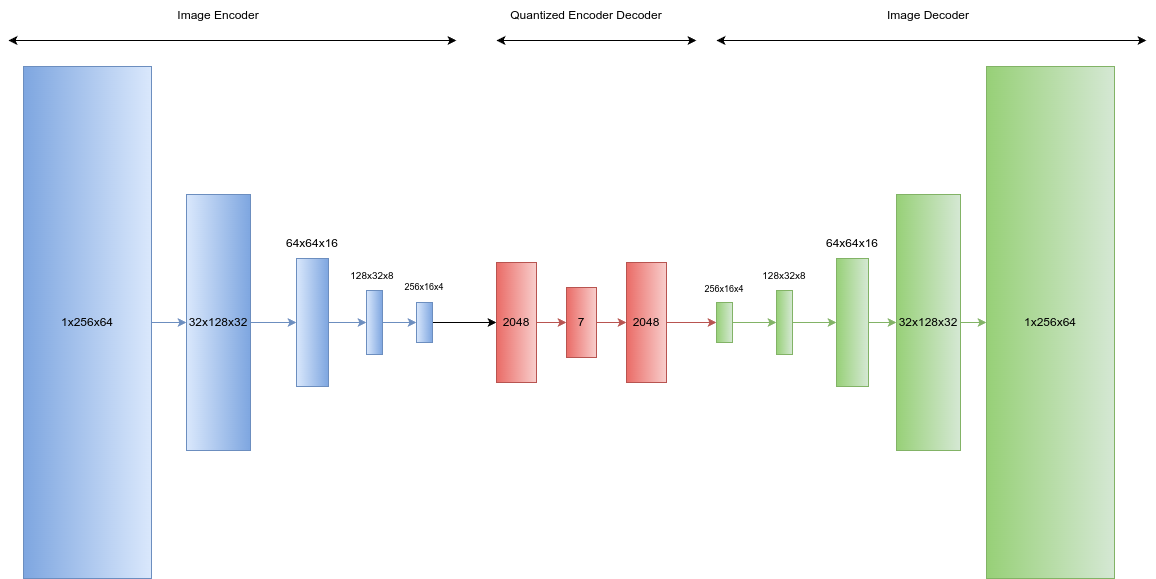}
  \caption{VQ-VAE architecture showing encoder, quantization module with codebook, and decoder.}
  \label{fig:VQVAE}
\end{figure}

\subsection{Generative Modeling of Latent Space}
Following VQ-VAE training \cite{13}, classical and quantum generative models were employed to model the prior distribution of the discrete latent space. Three models were implemented and evaluated: a QCBM \cite{15}, a QGAN \cite{16, 17}, and a classical LSTM network \cite{25} as baseline.

\subsubsection{Quantum Circuit Born Machine}
The QCBM implementation used the PennyLane library with GPU acceleration \cite{14, 15}. The approach involved training seven independent QCBM models, each dedicated to modeling the empirical distribution of a single dimension of the 7-dimensional latent space vector.

\begin{figure}[h]
  \centering
  \includegraphics[width=0.3\textwidth]{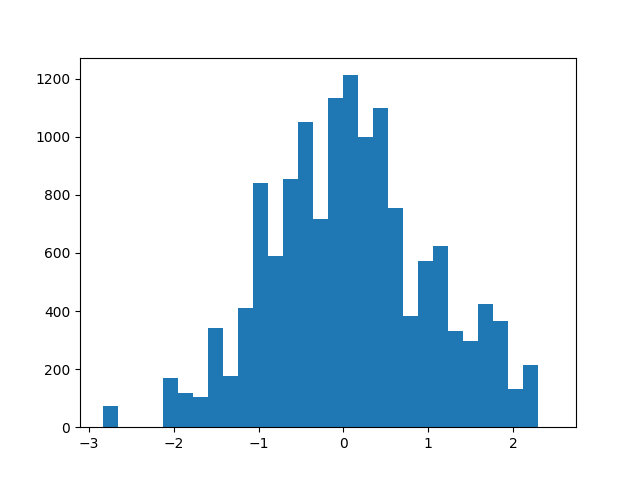}
  \caption{Distribution of continuous latent space values, showing approximately Gaussian characteristics.}
  \label{fig:LATENTSPACE}
\end{figure}

Based on the observed approximately Gaussian distribution of latent space values (Figure~\ref{fig:LATENTSPACE}), a custom Gaussian quantization function mapped values into 256 bins. Each bin value was then encoded into an 8-bit binary string, serving as the target output quantum state.

Each QCBM circuit used 8 qubits initialized in the $|0\rangle$ state, with 7 layers. Each layer comprised single-qubit $R_y$ rotation gates applied to all 8 qubits, followed by circular entanglement using Controlled-Z (CZ) gates \cite{14, 15}. The output yields a probability distribution over 256 possible computational basis states.

Training was performed using PennyLane's PyTorch interface and the lightning.gpu device. Circuit parameters were optimized using Adam optimizer with learning rate 0.1 over 100 iterations. The parameter-shift rule was employed for computing gradients \cite{15}. The loss function was the squared MMD between the QCBM distribution and the target empirical distribution \cite{27}, using an RBF kernel:
\begin{equation}
K(\mathbf{x}, \mathbf{x}') = \exp\left(-\frac{\|\mathbf{x} - \mathbf{x}'\|^2}{2\sigma^2}\right)
\end{equation}
with Gaussian bandwidths $\sigma \in \{0.25, 0.5, 1.0\}$ \cite{27}.

For sampling, a \texttt{QCBMSampler} class generates a full 7-dimensional latent vector by sampling independently from each of the seven trained QCBMs.

\begin{figure}[h]
  \centering
  \includegraphics[width=0.5\textwidth]{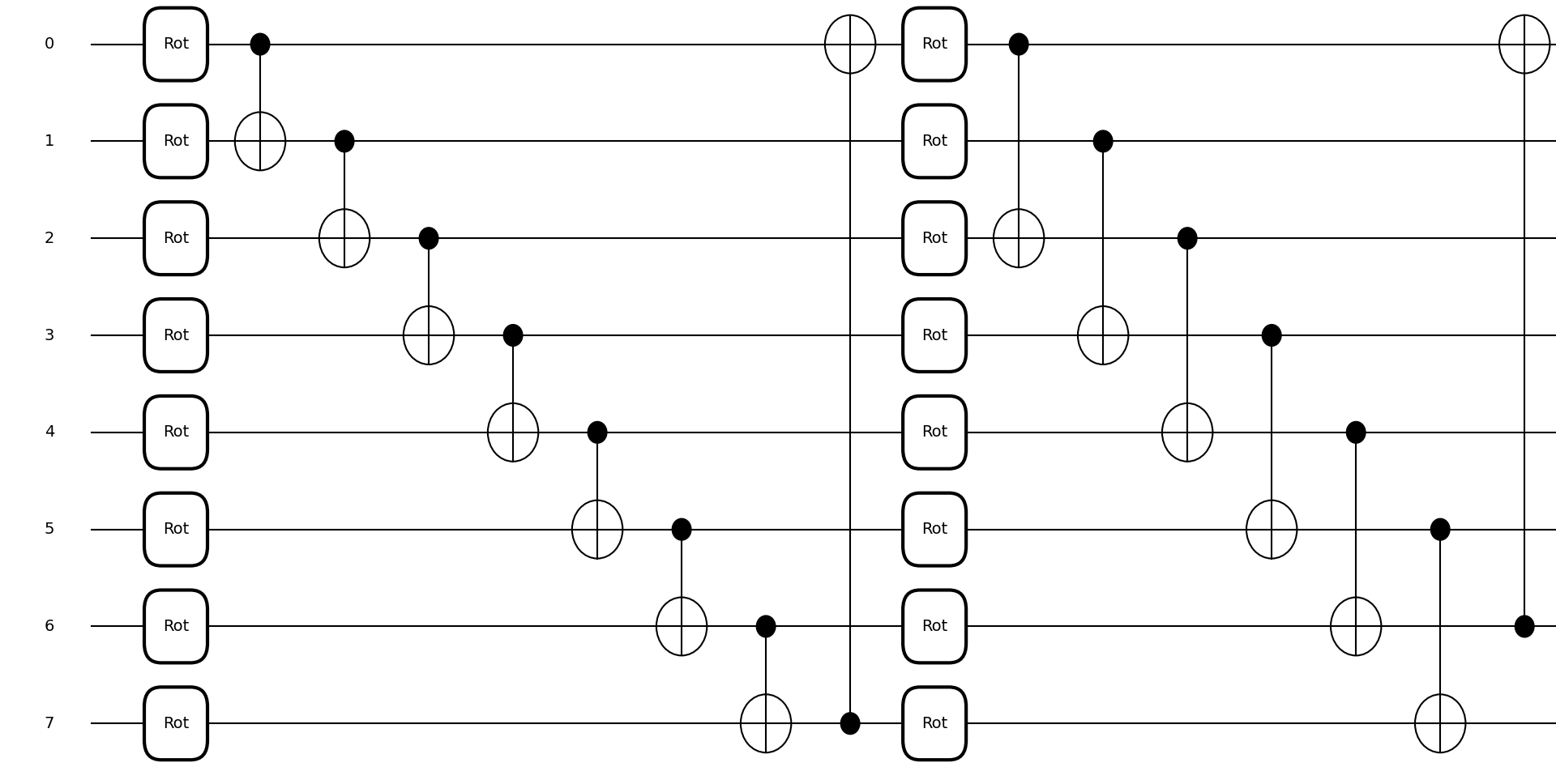}
  \caption{Two layers from the QCBM architecture showing $R_y$ rotation gates and CZ entanglement pattern.}
  \label{fig:QCBM}
\end{figure}

\begin{figure}[h]
  \centering
  \includegraphics[width=0.4\textwidth]{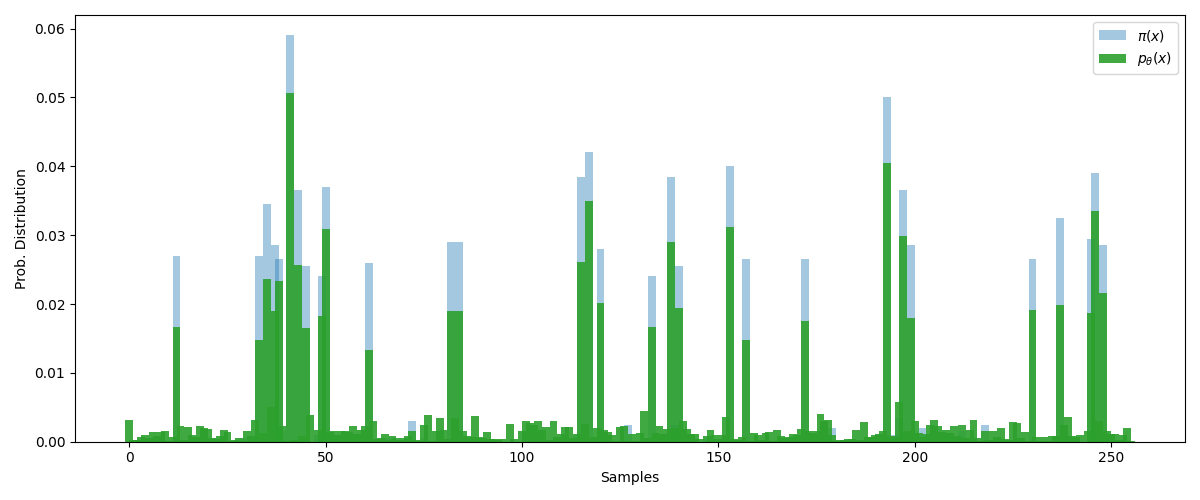}
  \includegraphics[width=0.4\textwidth]{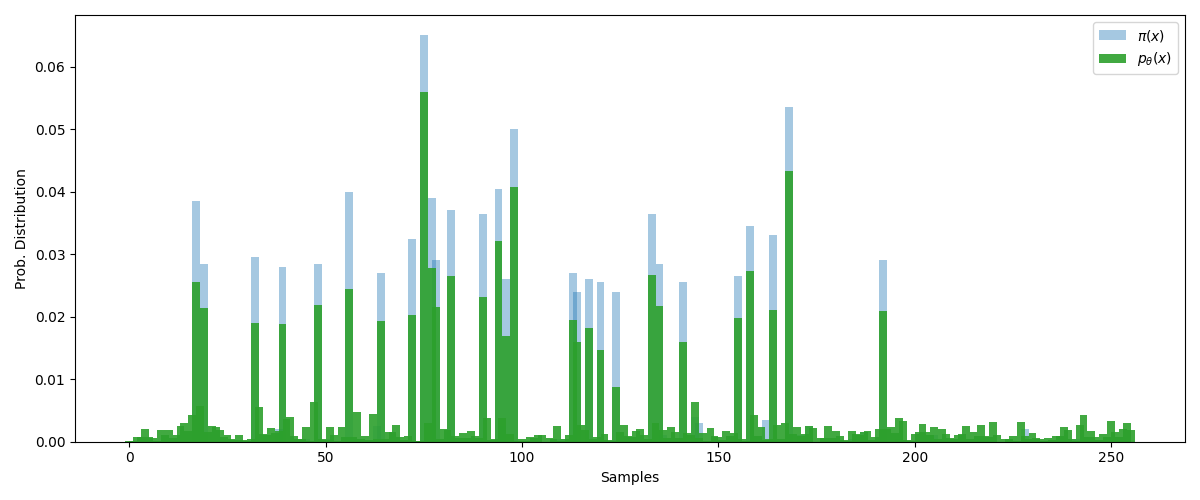}
  \caption{QCBM learned distributions (blue) compared to target distributions (green) for latent dimensions 1 and 7.}
  \label{fig:QCBM_distribution}
\end{figure}

\textbf{Independence Assumption.} Our QCBM approach models each latent dimension independently, effectively assuming $P(z_1, \ldots, z_7) \approx \prod_{i=1}^{7} P(z_i)$. This is a simplifying assumption that may not hold for general latent spaces. However, VQ-VAE training encourages the codebook to span the data manifold efficiently, which may induce approximate independence between dimensions. We did not formally test this assumption. The strong performance of QCBM relative to QGAN---which can capture dependencies through joint training---may suggest either that the latent dimensions are approximately independent for this dataset, or that the QGAN was undertrained relative to its capacity. Distinguishing these hypotheses is an important direction for future work.

\subsubsection{Quantum Generative Adversarial Network}
A QGAN was implemented to model the latent space distribution \cite{5, 16, 17}. The architecture consisted of a classical discriminator and a quantum generator.

The Discriminator was a classical feedforward neural network with three layers. It received a flattened input representing the probability distribution across 256 bins for each of the 7 latent dimensions ($256 \times 7$ matrix), mapped to a single scalar output with sigmoid activation, indicating the probability that the input distribution was real.

The Generator consisted of seven independent quantum circuits. Each generator used 10 qubits: 8 for the actual input (discretized latent values) and 2 ancillary qubits for entanglement. Each circuit consisted of 6 layers, with Pauli-$R_y$ rotation gates followed by full circular entanglement using CZ gates \cite{14, 16, 17}.

All seven generator outputs were combined to form a complete 7-dimensional latent vector. Training used separate Adam optimizers with learning rate 0.01, batch size 32, for 2 epochs. The discriminator minimized binary cross-entropy loss; the generator maximized the discriminator's output for generated samples \cite{5, 16, 17}.

\begin{figure}[h]
  \centering
  \includegraphics[width=0.4\textwidth]{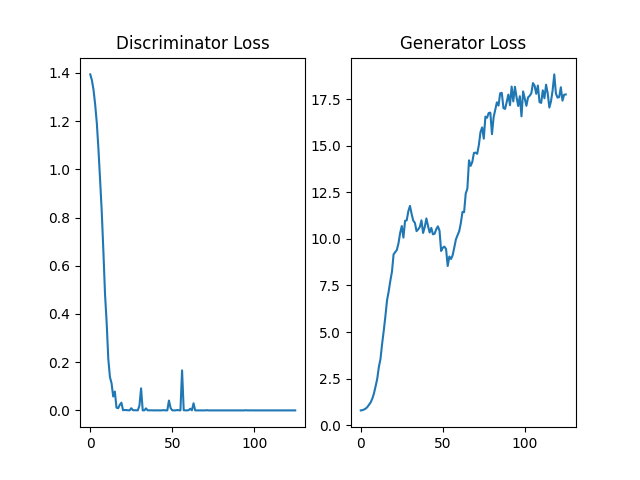}
  \caption{QGAN training losses showing discriminator and generator loss evolution.}
  \label{fig:QGANLOSSES}
\end{figure}

At inference time, random noise is bin-discretized and passed through the generator. The output is mapped back to continuous latent space using a reverse Gaussian bin-discretization function.

\subsubsection{Long Short-Term Memory}
As a classical baseline, an LSTM network was implemented to model the prior distribution \cite{25}. The latent space dimensions were treated as elements in a sequence.

The LSTM architecture consisted of a single layer with 256 neurons trained for 100 epochs. During training, the model received random noise as initial input and employed teacher forcing, where the true output from the previous time step was fed as input to the current step \cite{29, 30}. The training objective minimized the Mean Squared Error between predicted and true latent dimensions.

\begin{figure}[h]
  \centering
  \includegraphics[width=0.4\textwidth]{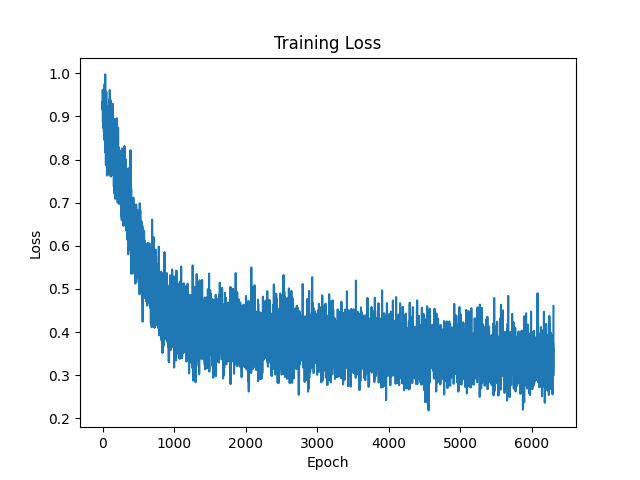}
  \caption{LSTM MSE training loss over iterations.}
  \label{fig:LSTMLOSSES}
\end{figure}

\textbf{Baseline Selection.} We selected LSTM for its natural fit to sequential generation tasks. We acknowledge that more sophisticated classical generative models exist (normalizing flows, diffusion models, transformers) and may perform differently. Our choice was motivated by: (1)~establishing a reference point for comparison, (2)~computational constraints during this exploratory phase, and (3)~focus on quantum model implementations as the primary technical contribution. Comparison against state-of-the-art classical generative models would strengthen future studies.

\subsection{Training Regime Considerations}
The three generative models were trained under different regimes due to their distinct architectures and convergence behaviors. Table~\ref{tab:training} summarizes key parameters.

\begin{table}[h]
\centering
\caption{Summary of Model Architectures and Training Parameters}
\label{tab:training}
\begin{tabular}{@{}lccc@{}}
\toprule
\textbf{Model} & \textbf{Architecture} & \textbf{Training} & \textbf{Loss} \\
\midrule
VQ-VAE & Conv.\ AE & 50 epochs & MSE + VQ \\
QCBM & 7$\times$(8q, 7L) & 100 iter. & MMD \\
QGAN & 7$\times$(10q, 6L) + MLP & 2 epochs & BCE/Adv. \\
LSTM & 1L, 256h & 100 epochs & MSE \\
\bottomrule
\end{tabular}
\begin{flushleft}
\small \textit{Note:} q = qubits, L = layers, h = hidden units, iter.\ = iterations.
\end{flushleft}
\end{table}

The QCBM models were each trained for 100 parameter updates. The QGAN was trained for 2 full passes through the dataset (approximately 125 parameter updates per generator). The LSTM was trained for 100 epochs.

We acknowledge that these training budgets are not directly comparable. The quantum models required substantially more wall-clock time per update due to simulation overhead, while the LSTM converged more quickly in terms of epochs but with different update dynamics. A rigorous comparison would require either equalizing computational budget or training all models to verified convergence with early stopping; we leave such analysis to future work. The results should be interpreted as observations under these specific experimental conditions.

\subsection{Model Comparison}
Comprehensive comparative analysis was conducted to evaluate the trained QCBM, QGAN, and LSTM models in capturing and sampling from the VQ-VAE's latent space distribution. Samples were generated from each trained model and compared against the original VQ-VAE codebook data as ground truth.

\subsubsection{t-SNE Visualization}
T-distributed Stochastic Neighbor Embedding (t-SNE) was used to visualize the high-dimensional latent spaces in two dimensions \cite{28}. A perplexity value of 100 was used. To ensure relative comparison, the t-SNE model was fitted on the combined dataset from all three generative models.

\subsubsection{PCA Visualization}
Principal Component Analysis (PCA) was also employed for dimensionality reduction \cite{8}. PCA was performed on the concatenated dataset from all models. Projection onto the first two principal components provides a linear perspective on the structure and spread of distributions, complementing the non-linear view from t-SNE.

\subsubsection{Average Minimum Distances}
Quantitative comparison was performed by calculating average minimum Euclidean distance between samples generated by each model and the original VQ-VAE codebook vectors \cite{26, 27}. For each generated sample, the minimum distance to any codebook vector was computed. The average across all generated samples provides a numerical measure of how closely the generated distribution approximates the true distribution.

\subsubsection{Nearest Neighbor Analysis}
Nearest Neighbor analysis provided a discretized evaluation \cite{26, 27}. For each codebook vector, the closest generated sample among all samples from all three models was identified. This determined which model's samples were most frequently the nearest neighbors to true latent space vectors, offering insight into which model best covers the high-density regions.

\subsubsection{Distance Distribution Analysis}
The minimal distance between each codebook vector and the closest sample generated by each model independently was computed. This allowed plotting and comparing the distance distributions, offering quantitative insight into each model's ability to reproduce the original latent space accurately.

\section{Results and Interpretation}
This section presents the experimental results and analysis. The findings are interpreted in the context of research objectives, with attention to experimental design limitations.

\subsection{LBM Fluid Simulator Performance}
The GPU-accelerated LBM simulator demonstrated robust performance and capability to generate physically accurate fluid dynamics data. Validation against established test cases confirmed the simulator's fidelity in reproducing expected flow phenomena, including vortex shedding at appropriate Reynolds numbers. The efficiency gained through GPU acceleration was critical for generating large datasets within reasonable timeframes. The primary dataset consisted of 2D vorticity profiles of flow around a cylinder at Re=500 and 0.1 Mach inlet velocity.

\subsection{VQ-VAE Latent Space Learning}
\begin{figure}[h]
  \centering
  \includegraphics[width=0.4\textwidth]{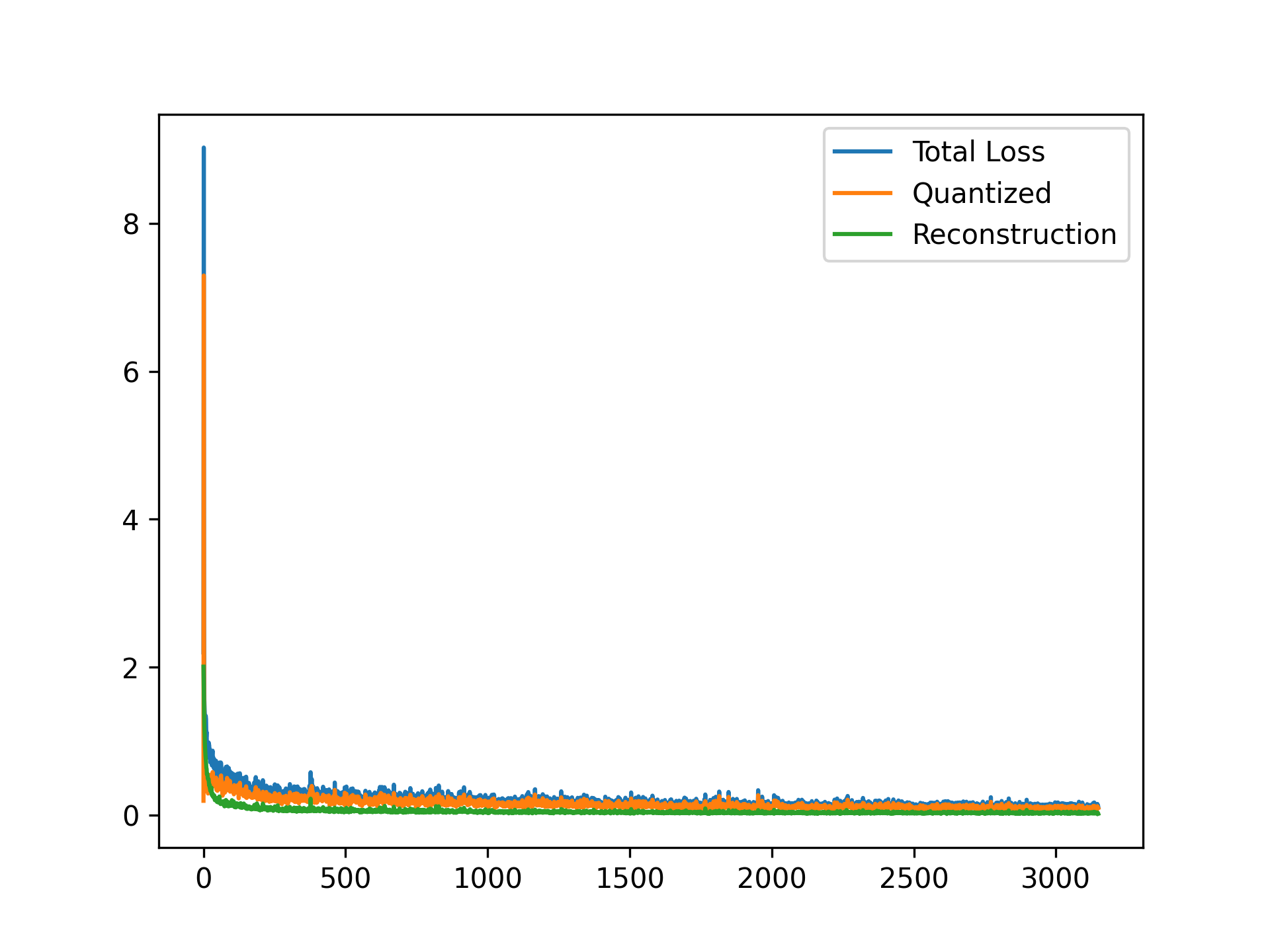}
  \caption{VQ-VAE training losses showing reconstruction, quantization, and total loss convergence.}
  \label{fig:VQVAELOSSES}
\end{figure}

The VQ-VAE was successfully trained to learn a compressed, discrete latent space representation \cite{13}. The training process exhibited stable convergence, as indicated by loss curves (Figure~\ref{fig:VQVAELOSSES}).

\begin{figure}[h]
  \centering
  \includegraphics[width=0.4\textwidth]{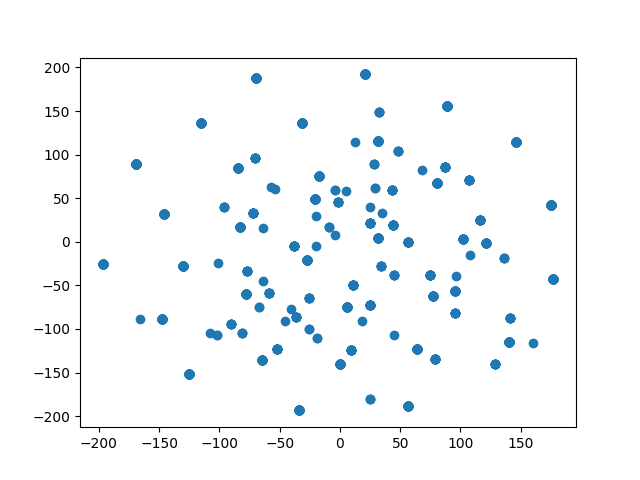}
  \caption{t-SNE visualization of the VQ-VAE codebook latent space.}
  \label{fig:VQVAESPACE}
\end{figure}

Analysis of the learned discrete latent space, visualized through t-SNE \cite{28}, revealed a structured distribution (Figure~\ref{fig:VQVAESPACE}). The projection showed a somewhat clustered distribution with discernible overall structure, suggesting successful dimensionality reduction while preserving meaningful variations in flow patterns.

\subsection{Comparative Analysis of Generative Models}
The primary objective was comparing classical and quantum generative models in learning and sampling from the discrete latent space. Analysis used both visualization techniques and quantitative metrics.

\subsubsection{Latent Space Visualization}
\begin{figure}[h]
  \centering
  \includegraphics[width=0.5\textwidth]{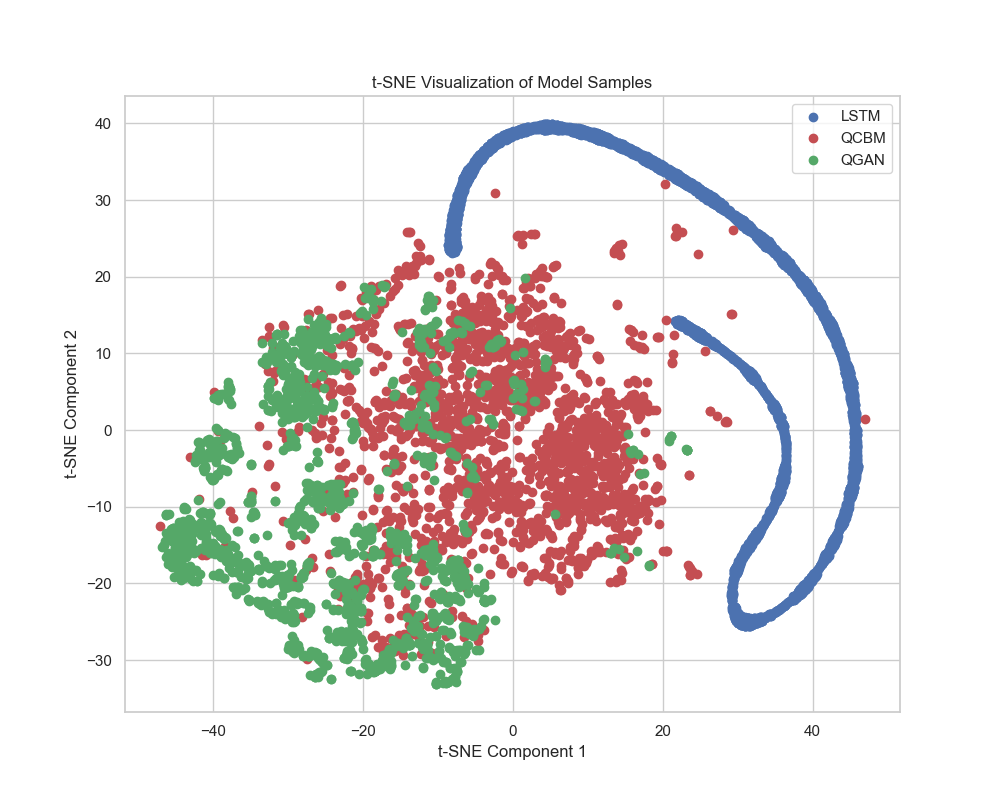}
  \caption{t-SNE comparative visualization of samples from LSTM, QCBM, and QGAN models.}
  \label{fig:TSNEVIS}
\end{figure}

Visualization using dimensionality reduction provided qualitative insights. The t-SNE projection (Figure~\ref{fig:TSNEVIS}) revealed distinct patterns.

The LSTM samples exhibited a relatively constrained, almost linear-like structure in the 2D projection space, suggesting the LSTM may have struggled to fully capture potentially non-linear correlations in the multi-dimensional latent space distribution.

In contrast, the quantum models produced samples occupying a broader region of the projected latent space, indicating potentially greater coverage of the latent distribution \cite{14}. The QCBM samples formed a more concentrated, Gaussian-like cluster in the center. The QGAN samples, while also covering a wider area than LSTM, appeared more fragmented and clustered in different regions.

\begin{figure}[h]
  \centering
  \includegraphics[width=0.51\textwidth]{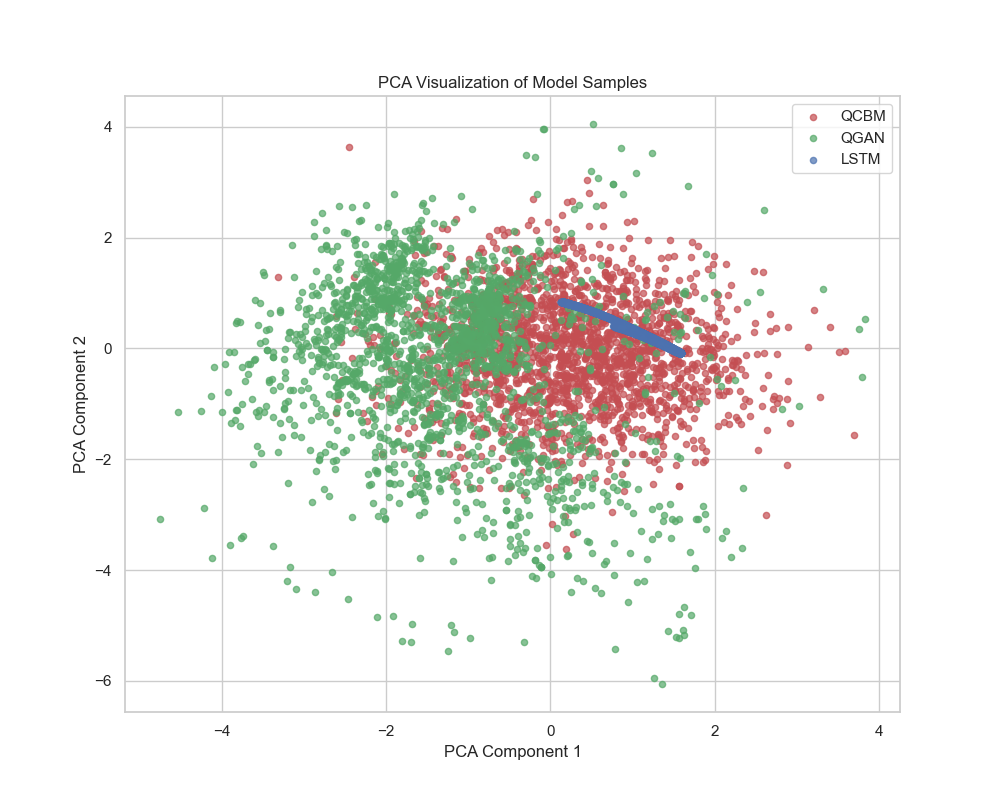}
  \caption{PCA comparative visualization showing first two principal components for each model's samples.}
  \label{fig:PCAVIS}
\end{figure}

The PCA visualization (Figure~\ref{fig:PCAVIS}) further supported these observations \cite{8}.

The PCA results showed LSTM samples aligning predominantly along a linear manifold. QCBM samples formed a more centralized cluster consistent with a distribution spanning multiple principal component directions. QGAN samples were more scattered, reinforcing the observation of a more fragmented distribution capture.

\subsubsection{Quantitative Performance Metrics}
\begin{figure}[h]
  \centering
  \includegraphics[width=0.24\textwidth]{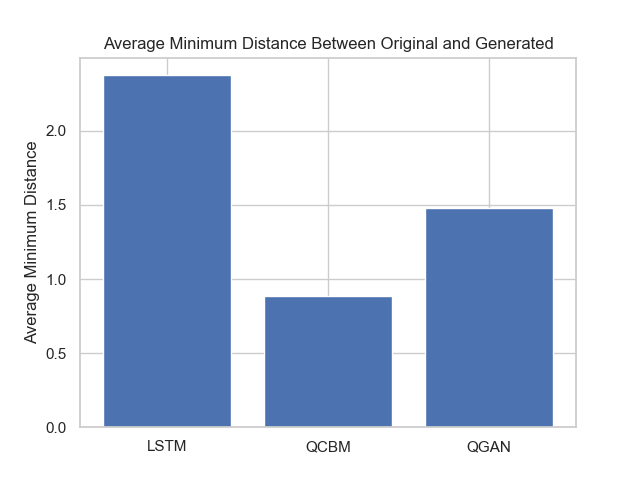} 
  \includegraphics[width=0.24\textwidth]{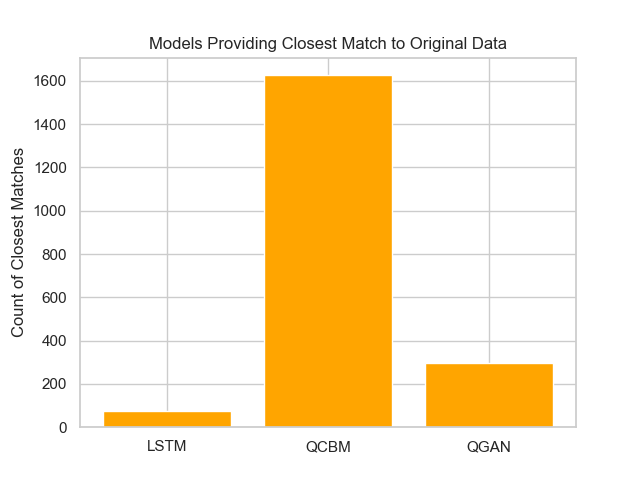} 
  \caption{Quantitative analysis: (Left) Average minimum distances to codebook for each model. (Right) Count of nearest neighbor matches per model.}
  \label{fig:QUANTAN}
\end{figure}

The average minimum Euclidean distance between generated samples and original codebook vectors served as a key performance indicator (Figure~\ref{fig:QUANTAN}).

Under our experimental conditions, the QCBM achieved the lowest average minimum distance (approximately 0.8), indicating its generated samples were, on average, closest to true latent space codewords. The QGAN had a higher average minimum distance (approximately 1.5) but lower than the LSTM. The LSTM exhibited the largest average minimum distance (approximately 2.4), consistent with visual assessment.

The Nearest Neighbor analysis revealed that QCBM samples were nearest neighbors to the vast majority of original codebook vectors (over 1600 out of 1999). QGAN had considerably fewer nearest neighbors (more than 200), and LSTM had very few samples closest to any codebook vector. This suggests QCBM was most effective at generating samples falling within the vicinity of actual data points.

\begin{figure}[h]
  \centering
  \includegraphics[width=0.5\textwidth]{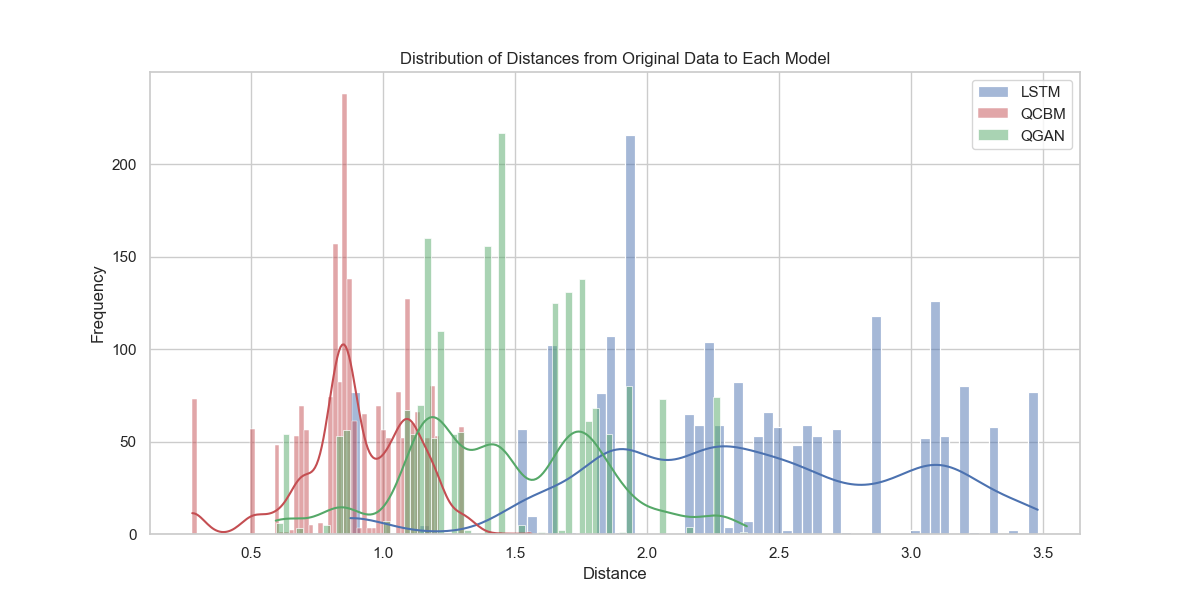}
  \caption{Distribution of minimum distances from original codebook vectors to closest generated samples for each model.}
  \label{fig:DISTDIST}
\end{figure}

Distance distribution histograms (Figure~\ref{fig:DISTDIST}) showed that QCBM had a sharp peak at a low distance value. The QGAN's distribution was broader and shifted towards higher distances, while LSTM's distribution was widely spread across much larger distances.

\subsection{Limitations of This Analysis}
Several factors limit the conclusions that can be drawn:

\begin{enumerate}
    \item \textit{Single experimental run}: All results represent single training runs without variance estimates. Stochastic training dynamics could produce different rankings with different random seeds.
    
    \item \textit{Unequal training budgets}: The models received different numbers of parameter updates and wall-clock training time. Observed performance differences may partly reflect training regime differences rather than fundamental model capabilities.
    
    \item \textit{Limited dataset}: The 1,999 vorticity snapshots represent a single flow configuration (cylinder, Re=500, 0.1 Mach). Generalization to different Reynolds numbers, geometries, or 3D simulations was not tested.
    
    \item \textit{Proxy metrics}: Average minimum distance and nearest neighbor counts measure proximity to training data but do not directly assess sample quality, diversity, or physical plausibility of decoded flow fields.
    
    \item \textit{Classical simulation}: All quantum circuits were simulated on classical hardware using PennyLane's lightning.gpu backend. This eliminates any computational advantage quantum hardware might provide and may introduce simulation-specific artifacts distinct from real quantum device behavior.
    
    \item \textit{Single baseline}: Comparison against only one classical model (LSTM) limits conclusions about quantum versus classical approaches more broadly.
\end{enumerate}

These limitations do not invalidate the observations but constrain their interpretation. We present these results as preliminary findings warranting further investigation rather than definitive conclusions.

\subsection{Summary of Observations}
The comparative analysis yielded the following observations under our experimental conditions:

\begin{itemize}
    \item The quantum generative models (QCBM and QGAN) produced samples with lower average minimum distances to the VQ-VAE codebook compared to the LSTM baseline.
    \item The QCBM, modeling each latent dimension independently, achieved the most favorable metrics among the models evaluated.
    \item The LSTM samples exhibited a constrained, linear-like structure in dimensionality-reduced visualizations, while quantum model samples showed broader coverage.
    \item The QGAN, despite being designed to capture dependencies, showed intermediate performance between QCBM and LSTM.
\end{itemize}

\section{Discussion}

\subsection{Possible Explanations for Observed Results}
Several hypotheses may explain the observed performance differences:

\textbf{Hypothesis 1: Architectural inductive biases.} Quantum circuits may have favorable inductive biases for modeling the specific distribution structure of this latent space. The quantum circuits' ability to represent probability distributions through Born rule sampling may naturally align with the approximately Gaussian marginal distributions of the VQ-VAE latent dimensions.

\textbf{Hypothesis 2: LSTM baseline limitations.} The LSTM, while appropriate for sequential prediction, may not be well-suited to the structure of this particular latent space. A 7-dimensional latent vector may not have strong sequential dependencies that an LSTM could exploit. More sophisticated classical baselines might perform differently.

\textbf{Hypothesis 3: Training regime effects.} The different training durations and convergence criteria may favor certain models. The QCBM's simpler per-dimension training may have allowed more effective optimization than the QGAN's complex adversarial dynamics or the LSTM's sequential prediction objective.

\textbf{Hypothesis 4: Latent space independence.} If VQ-VAE latent dimensions are approximately independent, the QCBM's independent modeling approach would be well-matched to the data, while the QGAN's capacity for modeling dependencies would be unnecessary overhead.

\textbf{Hypothesis 5: Metric sensitivity.} The chosen metrics may favor the type of distributions produced by certain models. Different metrics (e.g., Wasserstein distance, coverage metrics) might yield different relative rankings.

We cannot definitively distinguish among these hypotheses with the current experimental design. Future work should address each through targeted experiments.

\subsection{What This Study Does and Does Not Show}
This study \textbf{does} demonstrate:
\begin{itemize}
    \item A complete, functional pipeline for applying quantum generative models to CFD latent spaces---the first such application to learned latent representations
    \item That quantum models can produce samples from physics-derived discrete latent spaces
    \item Interesting differences in the characteristics of samples from different model types
    \item Specific experimental conditions under which QCBM produced favorable metrics
\end{itemize}

This study \textbf{does not} demonstrate:
\begin{itemize}
    \item Quantum advantage (all experiments used classical simulation)
    \item That quantum models are generally superior to classical models for this task
    \item That results would generalize to other flow configurations or latent space designs
    \item That observed metric differences translate to practically meaningful improvements in downstream tasks
\end{itemize}

\subsection{Implications for Future Research}
The results, while preliminary, suggest several directions for future investigation. The ability to model and sample from a compressed latent space of fluid flow data \cite{13} could open avenues for more efficient data augmentation, generation of synthetic training data for downstream tasks, and potentially accelerating design optimization loops. However, demonstrating such practical utility requires further work.

The observation that a relatively simple quantum circuit (QCBM with independent dimension modeling) achieved favorable metrics compared to both a more complex quantum model (QGAN) and a classical baseline (LSTM) suggests that model complexity does not necessarily correlate with performance on this task. This aligns with broader findings in machine learning that simpler models often generalize better when appropriately matched to problem structure.

\section{Limitations and Future Work}
Beyond the experimental limitations discussed in Section~V-D, several broader limitations must be acknowledged:

\begin{itemize}
    \item The quantum models were evaluated using classical simulations, which do not capture the nuances and potential noise of actual quantum hardware \cite{14}.
    \item The scalability of these quantum approaches to significantly larger latent spaces or higher-dimensional CFD problems on near-term quantum devices remains a challenge.
    \item The study focused on a specific 2D fluid flow scenario (cylinder at Re=500), and generalizability to other flow regimes, geometries, or 3D simulations needs investigation.
    \item The VQ-VAE latent space properties, including optimal number of dimensions and codebook size, could influence generative model performance \cite{13}.
\end{itemize}

Future research should address these limitations through:
\begin{itemize}
    \item \textbf{Rigorous benchmarking}: Equalize training budgets, perform multiple runs with different random seeds for variance estimates, and include stronger classical baselines (normalizing flows, diffusion models, VAE priors).
    \item \textbf{Physical validation}: Decode generated latent vectors and evaluate whether resulting flow fields satisfy physical constraints and appear qualitatively realistic.
    \item \textbf{Scalability analysis}: Test with larger latent dimensions and more qubits to understand scaling behavior.
    \item \textbf{Real quantum hardware}: Execute circuits on actual quantum devices to assess noise effects and practical feasibility.
    \item \textbf{Broader datasets}: Evaluate on multiple flow configurations, Reynolds numbers, and potentially 3D simulations.
    \item \textbf{Independence testing}: Formally analyze the correlation structure of the VQ-VAE latent space.
    \item \textbf{Downstream task evaluation}: Demonstrate practical utility by using generated samples for data augmentation or surrogate model training.
\end{itemize}

\section{Conclusion}
This research presents an exploratory study on the application of quantum generative models to compressed latent space representations of computational fluid dynamics data. While recent work has begun exploring quantum models for learning fluid statistics \cite{31}, the application to explicitly learned latent representations---particularly discrete codebook spaces from VQ-VAE architectures---has not been previously investigated. We developed a GPU-accelerated Lattice Boltzmann Method simulator \cite{18, 19} to generate fluid vorticity data, which was compressed into a discrete 7-dimensional latent space using a Vector Quantized Variational Autoencoder \cite{13}. This established infrastructure for comparing generative modeling approaches on physics-derived latent spaces.

We compared a classical Long Short-Term Memory network \cite{25} against two quantum generative models---the Quantum Circuit Born Machine \cite{15} and the Quantum Generative Adversarial Network \cite{16, 17}---for learning and sampling from this latent space distribution. Under our specific experimental conditions, the quantum models produced samples with lower average minimum distances to the true distribution compared to the LSTM baseline, with the QCBM achieving the most favorable metrics as measured by average minimum distance and nearest neighbor analysis, complemented by qualitative visualizations (t-SNE \cite{28}, PCA \cite{8}).

The primary contributions of this work are: (1)~a complete, open-source pipeline for quantum-classical comparison on physics simulation data; (2)~the first application of quantum generative models to learned latent space representations of CFD data, complementing recent work on quantum models for fluid statistics; and (3)~preliminary empirical observations and hypotheses that may guide future research. This work provides a foundation for more rigorous investigations at the intersection of quantum machine learning and computational physics.

All code is available at \url{https://github.com/AchrafHsain7/FluidX} to enable reproducibility and extension of this work.

\section*{Acknowledgment}
The author thanks Dr.\ Fouad Mohammed Abbou for guidance and discussions throughout the research. Dr.\ Lamiae Bouanane and Professor Mouna Kettani are acknowledged for their roles in supporting quantum computing research within the School of Science and Engineering. The conceptualization, technical development, and preparation of the manuscript were carried out by the first author.

\bibliographystyle{IEEEtran}

\begin{thebibliography}{31}

\bibitem{1}
M.~N.~Dhaubhadel, ``Review: CFD Applications in the Automotive Industry,'' \emph{J.\ Fluids Eng.}, vol.~118, no.~4, pp.~647--653, Dec.\ 1996.

\bibitem{2}
M.~Mani and A.~J.~Dorgan, ``A perspective on the state of aerospace CFD technology,'' \emph{Annu.\ Rev.\ Fluid Mech.}, vol.~55, pp.~431--457, Jan.\ 2023.

\bibitem{3}
R.~McConkey, E.~Yee, and F.-S.~Lien, ``A curated dataset for data-driven turbulence modelling,'' \emph{Sci.\ Data}, vol.~8, Art.\ 255, Aug.\ 2021.

\bibitem{4}
S.~Bond-Taylor \emph{et al.}, ``Deep generative modelling: A comparative review of VAEs, GANs, normalizing flows, energy-based and autoregressive models,'' \emph{IEEE Trans.\ Pattern Anal.\ Mach.\ Intell.}, vol.~44, no.~11, pp.~7327--7347, Nov.\ 2021.

\bibitem{5}
I.~J.~Goodfellow \emph{et al.}, ``Generative adversarial nets,'' in \emph{Adv.\ Neural Inf.\ Process.\ Syst.}, vol.~27, 2014.

\bibitem{6}
D.~P.~Kingma and M.~Welling, ``Auto-encoding variational bayes,'' in \emph{Proc.\ Int.\ Conf.\ Learn.\ Represent.}, 2014.

\bibitem{7}
P.~Baldi, ``Autoencoders, unsupervised learning, and deep architectures,'' in \emph{Proc.\ ICML Workshop Unsupervised Transfer Learn.}, JMLR Workshop Conf.\ Proc., 2012.

\bibitem{8}
G.~E.~Hinton and R.~R.~Salakhutdinov, ``Reducing the dimensionality of data with neural networks,'' \emph{Science}, vol.~313, no.~5786, pp.~504--507, Jul.\ 2006.

\bibitem{9}
P.~Vincent \emph{et al.}, ``Stacked denoising autoencoders: Learning useful representations in a deep network with a local denoising criterion,'' \emph{J.\ Mach.\ Learn.\ Res.}, vol.~11, no.~12, 2010.

\bibitem{11}
D.~J.~Rezende, S.~Mohamed, and D.~Wierstra, ``Stochastic backpropagation and approximate inference in deep generative models,'' in \emph{Proc.\ Int.\ Conf.\ Mach.\ Learn.}, 2014.

\bibitem{12}
I.~Higgins \emph{et al.}, ``$\beta$-VAE: Learning basic visual concepts with a constrained variational framework,'' in \emph{Proc.\ Int.\ Conf.\ Learn.\ Represent.}, 2017.

\bibitem{13}
A.~van~den~Oord and O.~Vinyals, ``Neural discrete representation learning,'' in \emph{Adv.\ Neural Inf.\ Process.\ Syst.}, vol.~30, 2017.

\bibitem{14}
M.~Hibat-Allah, M.~Mauri, J.~Carrasquilla, and A.~Perdomo-Ortiz, ``A framework for demonstrating practical quantum advantage: comparing quantum against classical generative models,'' \emph{Commun.\ Phys.}, vol.~7, pp.~1--9, 2024, doi:10.1038/s42005-024-01552-6.

\bibitem{15}
M.~Benedetti \emph{et al.}, ``A generative modeling approach for benchmarking and training shallow quantum circuits,'' \emph{NPJ Quantum Inf.}, vol.~5, no.~1, p.~45, 2019.

\bibitem{16}
S.~Lloyd and C.~Weedbrook, ``Quantum generative adversarial learning,'' \emph{Phys.\ Rev.\ Lett.}, vol.~121, no.~4, p.~040502, Jul.\ 2018.

\bibitem{17}
P.-L.~Dallaire-Demers and N.~Killoran, ``Quantum generative adversarial networks,'' \emph{Phys.\ Rev.\ A}, vol.~98, no.~1, p.~012324, Jul.\ 2018.

\bibitem{18}
S.~Chen and G.~D.~Doolen, ``Lattice Boltzmann method for fluid flows,'' \emph{Annu.\ Rev.\ Fluid Mech.}, vol.~30, no.~1, pp.~329--364, Jan.\ 1998.

\bibitem{19}
J.~Latt \emph{et al.}, ``Multi-GPU Programming with Standard Parallel C++, Part 2,'' \emph{NVIDIA Developer Blog}, Apr.\ 18, 2022. [Online]. Available: https://developer.nvidia.com/blog/multi-gpu-programming-with-standard-parallel-cpp-part-2/

\bibitem{20}
Q.~Zou and X.~He, ``On pressure and velocity boundary conditions for the lattice Boltzmann BGK model,'' \emph{Phys.\ Fluids}, vol.~9, no.~6, pp.~1591--1598, Jun.\ 1997.

\bibitem{21}
M.~Bouzidi, M.~Firdaouss, and P.~Lallemand, ``Momentum transfer of a Boltzmann-lattice fluid with boundaries,'' \emph{Phys.\ Fluids}, vol.~13, no.~11, pp.~3452--3459, Nov.\ 2001.

\bibitem{22}
C.~L.~M.~H.~Navier, ``M\'{e}moire sur les lois du mouvement des fluides,'' \emph{M\'{e}m.\ R.\ Acad.\ Sci.\ Inst.\ France}, vol.~6, pp.~389--440, 1823.

\bibitem{23}
G.~G.~Stokes, ``On the theories of the internal friction of fluids in motion, and of the equilibrium and motion of elastic solids,'' \emph{Trans.\ Cambridge Philos.\ Soc.}, vol.~8, pp.~287--319, 1845.

\bibitem{24}
O.~Reynolds, ``An Experimental Investigation of the Circumstances which Determine whether the Motion of Water shall be Direct or Sinuous,'' \emph{Phil.\ Trans.\ R.\ Soc.\ Lond.}, vol.~174, pp.~935--982, 1883.

\bibitem{25}
S.~Hochreiter and J.~Schmidhuber, ``Long short-term memory,'' \emph{Neural Comput.}, vol.~9, no.~8, pp.~1735--1780, Nov.\ 1997.

\bibitem{26}
D.~M.~Blei, A.~Kucukelbir, and J.~D.~McAuliffe, ``Variational inference: A review for statisticians,'' \emph{J.\ Am.\ Stat.\ Assoc.}, vol.~112, no.~518, pp.~859--877, 2017.

\bibitem{27}
A.~Gretton \emph{et al.}, ``A kernel two-sample test,'' \emph{J.\ Mach.\ Learn.\ Res.}, vol.~13, no.~1, pp.~723--773, Mar.\ 2012.

\bibitem{28}
L.~van~der~Maaten and G.~Hinton, ``Visualizing data using t-SNE,'' \emph{J.\ Mach.\ Learn.\ Res.}, vol.~9, no.~11, pp.~2579--2605, 2008.

\bibitem{29}
A.~M.~Lamb \emph{et al.}, ``Professor forcing: A new algorithm for training recurrent networks,'' in \emph{Adv.\ Neural Inf.\ Process.\ Syst.}, vol.~29, 2016.

\bibitem{30}
Y.~Bengio \emph{et al.}, ``Scheduled sampling for sequence prediction with recurrent neural networks,'' in \emph{Adv.\ Neural Inf.\ Process.\ Syst.}, vol.~28, 2015.

\bibitem{31}
A.~Zlokapa \emph{et al.}, ``Quantum-informed machine learning for predicting spatiotemporal chaos,'' arXiv preprint arXiv:2507.19861, 2025.

\end{thebibliography}

\end{document}